\documentclass[runningheads]{llncs}
\usepackage[T1]{fontenc}
\usepackage{graphicx}
\usepackage{multicol}
\usepackage{multirow}
\usepackage{comment}
\usepackage{algorithm}
\usepackage{algpseudocode}
\usepackage{amsmath}
\usepackage{amsfonts}
\usepackage{hyperref}
\usepackage{xcolor}

\usepackage{booktabs}
\usepackage{float}

\begin{document}
\title{Clickbait detection: quick inference with maximum impact}
\titlerunning{Hybrid Clickbait Detection}
%
\author{
Soveatin Kuntur\inst{1}\orcidID{0000-0001-6029-941X} \and
Panggih Kusuma Ningrum\inst{2}\orcidID{0000-0002-8630-6603} \and
Anna Wróblewska\inst{1}\orcidID{0000-0002-3407-7570} \and
Maria Ganzha\inst{1}\orcidID{0000-0001-7714-4844} \and
Marcin Paprzycki\inst{3}\orcidID{0000-0002-8069-2152}
}

\authorrunning{S. Kuntur et al.}

\institute{
Faculty of Mathematics and Information Science, Warsaw University of Technology, plac Politechniki 1, 00-661 Warsaw, Poland \and
Université Marie et Louis Pasteur, CRIT, F-25000 Besançon, France \and
 Systems Research Institute, Polish Academy of Sciences, Newelska 6, 01-447 Warsaw, Poland
}

\maketitle              
\begin{abstract}
We propose a lightweight hybrid approach to clickbait detection that combines OpenAI semantic embeddings with six compact heuristic features capturing stylistic and informational cues. To improve efficiency, embeddings are reduced using PCA and evaluated with XGBoost, GraphSAGE, and GCN classifiers. While the simplified feature design yields slightly lower F1-scores, graph-based models achieve competitive performance with substantially reduced inference time. High ROC--AUC values further indicate strong discrimination capability, supporting reliable detection of clickbait headlines under varying decision thresholds.
\end{abstract}


\section{Introduction}
Clickbait refers to online content specifically designed to entice readers to click a link while offering little substantive value in return \cite{Scott2021Clickbait} (see Table \ref{tab:headline_examples}). Although clickbait predates modern digital journalism and was once considered a relatively benign form of sensationalism \cite{10.1145/3309699}, this characterization has become increasingly outdated. As journalism has shifted toward digital platforms, clickbait has evolved into a pervasive mechanism for shaping user attention and engagement, often blurring the boundary between legitimate news and misleading presentation \cite{KhawarBoukes2025Sensationalism}. Early forms of clickbait were commonly associated with sensational topics such as crime, disasters, or human-interest stories \cite{adams1978local}, whereas more recent manifestations frequently exploit information gaps, emotional triggers, and rage-inducing phrasing \cite{shin2025emotion}.

\begin{table}[!htbp]
\centering
\small
\begin{tabular}{p{0.46\linewidth} |p{0.46\linewidth}}
\hline
\textbf{Clickbait headline} & \textbf{Non-clickbait headline} \\
\hline
``6 Things Most Doctors Won't Tell You About Dieting''

\vspace{0.6em}

``This Simple Investment Trick Could Make You Rich Overnight''

\vspace{0.6em}

&
``Aryna Sabalenka survives the Wimbledon heat: World No. 1 into semi-finals with rollercoaster comeback to beat Laura Siegemund, 37, despite 23C temperatures''

\\
\hline
\end{tabular}
\vspace{6pt}
\caption{Examples of clickbait and non-clickbait headlines. Clickbait headlines are typically short and designed to induce a curiosity gap, whereas non-clickbait headlines tend to be longer and provide explicit informational content.}
\label{tab:headline_examples}
\end{table}

From an economic perspective, clickbait is tightly coupled with profit-driven incentives, as increased click-through rates directly translate into higher advertising revenue and web traffic \cite{ApresjanOrlov2022Clickbait}. This incentive structure encourages the continued use of misleading headlines, even in the presence of moderation mechanisms and growing user awareness. Consequently, clickbait remains a persistent phenomenon across online news ecosystems.

Recent user-centered studies indicate that clickbait primarily exerts its influence at an early stage of content exposure, shaping user perceptions before the article body is accessed \cite{shrestha2025luring}. Since many users rely almost exclusively on headlines when forming initial judgments, effective clickbait detection must operate under limited contextual information. This requirement motivates headline-centric detection approaches that prioritize efficiency and robustness over deep contextual modeling.

\textcolor{black}{A wide range of machine learning approaches has been proposed for clickbait detection, including methods based on lexical features, pretrained word embeddings, and large language models \cite{wang2025clickbait}, \cite{alarfaj2025clickbait}} While these approaches often achieve strong predictive performance, many rely on full article content, multimodal inputs, or computationally intensive architectures, limiting their suitability for real-world deployment \cite{wang2025multi}, \cite{abdullah2026multimodal}. In contrast, practical systems must operate in constrained settings where only short text inputs -- such as headlines -- are available, and low-latency inference is essential.

Building on our prior dataset analysis \cite{kuntur2026fake}, we find that incorporating stylistic cues is essential for effective clickbait detection. Our benchmarking results indicate that, although Graph Neural Networks do not achieve the same accuracy as Transformer-based models, they offer faster inference times \cite{kuntur2024comparative}. In contrast, traditional methods such as TF-IDF are increasingly less relevant in the LLM era. Motivated by these findings, we propose a hybrid approach. Prior work (e.g., Chakraborty et al.~\cite{7752207stopclickbait}) demonstrates that combining semantic representations with handcrafted linguistic features improves performance in clickbait detection. However, existing studies largely focus on accuracy while overlooking computational efficiency. To address this, we examine the trade-off between accuracy and efficiency by applying PCA for dimensionality reduction and evaluating XGBoost \cite{xgboost10.1145/2939672.2939785}, GraphSAGE \cite{hamilton2018inductiverepresentationlearninglarge}, and GCN \cite{kipf2017semisupervisedclassificationgraphconvolutional}, with explicit measurement of inference latency.

\section{Methodology}

\subsection{Problem formulation and datasets}
We formulate clickbait detection as a binary classification task operating solely on article headlines. Given a headline $T$, the goal is to predict whether it exhibits clickbait characteristics ($y = 1$) or not ($y = 0$). This setting reflects early-stage detection scenarios, where access to full article content or multimodal information is unavailable, and low-latency inference is essential.
We construct our dataset by combining three publicly available sources: Kaggle-1 \cite{kaggle_1_source}, Kaggle-2 \cite{kaggle_2_source}, and Clickbait Challenge 2017 (CC17) \cite{clickbait_challenge_2017}. The Kaggle datasets provide article titles annotated with binary clickbait labels, while CC17 includes richer annotations that we reduce to a single clickbait label to ensure consistency. The resulting datasets are merged into a unified corpus. From this corpus, we randomly sample 20,000 clickbait and 20,000 non-clickbait instances across the three datasets to construct a balanced dataset.


\subsection{Feature representation}

\paragraph{Semantic embeddings.}
Semantic information is encoded using \href{https://developers.openai.com/api/docs/models/text-embedding-3-large}
{\texttt{openai\allowbreak-\allowbreak text\allowbreak-\allowbreak embedding\allowbreak-\allowbreak 3\allowbreak-\allowbreak large}}, which produces dense vector representations of headlines in a 3{,}072-dimensional space. To reduce computational cost while preserving discriminative capacity, we apply Principal Component Analysis (PCA) to project embeddings into a 1{,}000-dimensional subspace. This dimensionality reduction significantly lowers memory usage and accelerates downstream learning without noticeably degrading performance. Assuming standard 32-bit floating point storage \cite{numpy_float32}, PCA reduces memory usage from approximately 491~MB to 153~MB.

\paragraph{Heuristic scores.}
To complement semantic embeddings, we compute two scalar heuristic features designed to capture non-semantic properties commonly associated with clickbait language: (i) \textbf{Baitness score.} This measure quantifies the degree to which a headline employs attention-grabbing and curiosity-inducing cues. It aggregates normalized indicators, including punctuation patterns, capitalization, numerical expressions, sentiment intensity, readability, and the presence of predefined bait phrases following our previous work \cite{michaluk2026click}. (ii) \textbf{Informativeness score.} This measure captures factual density and specificity. It is computed using lexical density, numerical content, title length, and a penalty for vague or generic expressions that obscure concrete information.

Both scores are normalized to the range $[0,1]$ and designed to be computationally inexpensive, interpretable, and independent of complex linguistic preprocessing.

\paragraph{Hybrid feature vector.}
The final representation is obtained by concatenating the PCA-reduced semantic embedding with the baitness and informativeness scores, yielding a compact hybrid feature vector:
\[
\mathbf{x} = [\mathbf{e}_{\text{PCA}} \, \| \, b \, \| \, i],
\]
where $\mathbf{e}_{\text{PCA}} \in \mathbb{R}^{1000}$ denotes the reduced embedding, and $b$ and $i$ denote the baitness and informativeness scores, respectively.

\paragraph{Illustrative example.}
To illustrate the hybrid feature representation, Table~\ref{tab:feature_example} shows two example headlines and their corresponding heuristic scores. While semantic embeddings encode the overall meaning of the headline, the baitness and informativeness scores capture complementary stylistic and informational cues.

\begin{table}[H]
\centering
\small
\begin{tabular}{p{5.5cm} c c}
\toprule
\textbf{Headline} & \textbf{Baitness} & \textbf{Informativeness} \\
\midrule
``You Won’t Believe What This Celebrity Did!'' & 0.82 & 0.18 \\
``Study Finds 23\% Increase in Solar Adoption Across Europe'' & 0.21 & 0.79 \\
\bottomrule
\end{tabular}
\vspace{6pt}
\caption{Illustrative examples of baitness and informativeness scores. Clickbait headlines exhibit high baitness and low informativeness, while factual headlines show the opposite pattern.}
\label{tab:feature_example}
\end{table}
As shown in Table~\ref{tab:feature_example}, a provocative celebrity headline has high baitness (0.82) and low informativeness (0.18), while a science headline shows the opposite pattern. This demonstrates how heuristic scores complement semantic embeddings in distinguishing clickbait from legitimate content.


\subsection{Classifier Architectures}

To evaluate the impact of classifier choice on both predictive performance and efficiency, we apply three classification models to the same hybrid feature representation: (i) \textbf{XGBoost} \cite{xgboost10.1145/2939672.2939785}, a strong tree-based baseline commonly used in hybrid clickbait detection. (ii) \textbf{GraphSAGE} \cite{hamilton2018inductiverepresentationlearninglarge}, a graph neural network model applied to a similarity-based $k$-nearest neighbor graph constructed over the hybrid feature space. (iii) \textbf{GCN} \cite{kipf2017semisupervisedclassificationgraphconvolutional}, a graph convolutional network prioritizing low-latency inference. By holding the feature representation constant, this comparison isolates the effect of classifier architecture on accuracy and runtime characteristics. These classifiers were selected based on prior work \cite{kuntur2024comparative}. In our comparative analysis, we observed that Graph Neural Networks (GNNs) provide faster and promising results. Additionally, in a separate study \cite{michaluk2026click}, we found that XGBoost achieves strong performance, particularly when combined with feature extraction techniques. Building on these insights, our goal is to enable rapid inference while maintaining high impact and performance.


\subsection{Design Rationale}

Our design is guided by a “quick inference with maximum impact” principle. Instead of increasing feature complexity or relying on heavyweight linguistic pipelines, we focus on a small number of carefully designed heuristic signals that complement semantic embeddings. This approach reduces implementation complexity, improves reproducibility, and enables efficient inference, making it suitable for large-scale or real-time clickbait detection systems.

\section{Results and Discussion}

Table~\ref{tab:model_summary} reports the performance of the evaluated clickbait detection models using F1-score as the primary evaluation metric, alongside ROC--AUC and per-sample inference time. Per-sample inference time is computed by dividing the total evaluation time on the test set by the number of test instances, and reflects only the model forward pass during evaluation. The F1-score is emphasized, as it provides a balanced measure of precision and recall and is particularly suitable for clickbait detection, where both false positives and false negatives are undesirable. Across all classifiers, the hybrid feature representation combining OpenAI semantic embeddings with baitness and informativeness scores consistently outperforms embedding-only baselines. This confirms that stylistic and informational cues complement semantic representations, thereby improving class separability in headline-based clickbait detection.

Among the evaluated models, the GraphSAGE classifier achieves the highest F1-score (0.8572), indicating the best balance between precision and recall. This suggests that GraphSAGE is particularly effective at identifying clickbait headlines while avoiding excessive false positives. XGBoost attains a slightly lower F1-score (0.8465), demonstrating strong predictive capability but with inferior efficiency. The GCN model achieves an F1-score of 0.8382, reflecting a modest reduction in predictive performance compared to GraphSAGE.

\begin{figure}[h!]
\centering
\includegraphics[width=0.5\linewidth]{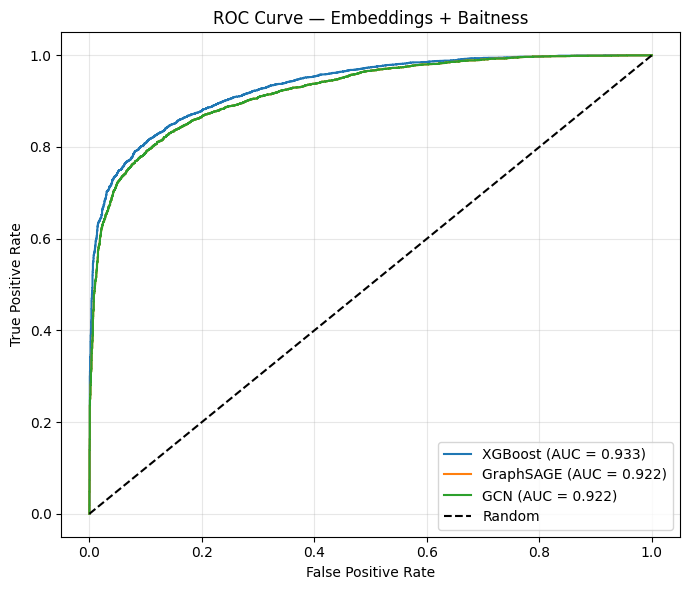}
\caption{ROC curves for hybrid clickbait detection models using semantic embeddings and baitness features. Graph-based models achieve competitive discrimination performance compared to XGBoost.}
\label{fig:roc_auc}
\end{figure}

Inference time analysis reveals a clear efficiency advantage for graph-based models. GraphSAGE reduces inference latency by approximately 25\% compared to XGBoost, while simultaneously improving the F1-score. GCN further prioritizes efficiency, achieving the lowest inference time (98.47~ms over 50 training epochs), at the cost of a small decrease in F1-score. This trade-off highlights the suitability of different classifiers for distinct deployment scenarios. Although F1-score is the primary evaluation metric, ROC--AUC values are additionally reported to assess ranking quality across decision thresholds. As illustrated in Figure~\ref{fig:roc_auc}, the high ROC--AUC scores achieved by GraphSAGE (0.9356) and XGBoost (0.9330) indicate strong discrimination capability, supporting the F1-based findings. In particular, elevated ROC--AUC values suggest reliable detection of true clickbait instances across varying classification thresholds. Overall, the results demonstrate that GraphSAGE offers the best balance between predictive performance and inference efficiency, while GCN is a compelling option for latency-sensitive applications. These findings underscore the importance of jointly considering accuracy and computational cost when designing practical clickbait detection systems.

\begin{table}[t]
\centering
\scriptsize
\begin{tabular}{p{6.6cm} r c c c}
\toprule
Feature Representation & Classifier & F1-score & ROC-AUC & Inference \\
 &  &  &  &  time (ms) \\
\midrule

\textit{Our proposed hybrid approach} & & & & \\
OpenAI Emb. (1000d) + Baitness + Informativeness  
& XGBoost & 0.8465 & 0.9330 & 236.65 \\
OpenAI Emb. (1000d) + Baitness + Informativeness  
& GraphSAGE & \textbf{0.8572} & \textbf{0.9356} & 177.79 \\
OpenAI Emb. (1000d) + Baitness + Informativeness  
& GCN & 0.8382 & 0.9219 & \textbf{98.47} \\
\bottomrule
\end{tabular}
\vspace{3pt}
\caption{Performance and inference time comparison of clickbait detection models. Inference time is reported per sample.}
\label{tab:model_summary}
\end{table}

\textcolor{black}{\section{Conclusion}}
\label{sec:conclusion}

This work investigated the trade-off between predictive performance and computational efficiency in hybrid clickbait detection under a deliberately lightweight feature design. Instead of relying on extensive handcrafted feature engineering, we combined deep semantic embeddings with only six compact heuristic measures capturing key stylistic and informational properties of headlines. This design choice reflects a minimum-inference time, maximum-impact perspective aimed at improving practical deployability.

While the reduced feature set yields slightly lower F1-scores than feature-heavy hybrid approaches, the consistently high ROC--AUC values indicate strong discrimination and reliable detection of true clickbait instances across decision thresholds. This suggests that the proposed representation preserves essential ranking information despite its simplicity, thereby boosting confidence in its practical effectiveness.

Comparative evaluation across classifiers further highlights the importance of architectural choice. Although XGBoost achieves competitive predictive performance, graph-based models, particularly GraphSAGE, offer a more favorable balance between accuracy and inference efficiency. The substantial reduction in inference time achieved by GraphSAGE demonstrates that meaningful performance can be retained while significantly lowering computational cost, making the approach suitable for high-throughput and latency-sensitive applications. \textbf{Embedding generation remains the dominant cost,} requiring 12{,}863.44 seconds to encode 40{,}000 headlines.

Overall, these findings emphasize that effective clickbait detection does not necessarily require extensive feature engineering or complex models. Instead, carefully selected heuristic signals, combined with semantic embeddings and efficient classifiers, can achieve robust performance while improving scalability. Future work will focus on leveraging our results as a backbone for developing a Chrome extension, with the aim of making this approach more practical and accessible at the end-user level.

\section*{Acknowledgments}
A.W and S.K were funded by the European Union under the Horizon Europe grant OMINO (grant number 101086321). Views and opinions expressed are, however, those of the authors
only and do not necessarily reflect those of the European Union or the European Research Executive Agency. Neither the European Union nor the European Research Executive Agency can be held responsible for them. A.W. and S.K. were also co-financed with funds from the Polish Ministry of Education and Science under the program entitled International Co-Financed Projects. 

 \bibliographystyle{splncs04}
 \bibliography{mybib}

\end{document}